\documentclass[conference]{IEEEtran}
\author{\IEEEauthorblockN{Anonymous Authors}} 
\IEEEoverridecommandlockouts
\usepackage{cite}
\usepackage{amsmath,amssymb,amsfonts}
\usepackage{algorithmic}
\usepackage{graphicx}
\usepackage{textcomp}
\usepackage{xcolor}
\usepackage{url}
\usepackage{hyperref}
\usepackage{tabularray}
\usepackage{multirow}
\usepackage{enumitem}
\def\BibTeX{{\rm B\kern-.05em{\sc i\kern-.025em b}\kern-.08em
    T\kern-.1667em\lower.7ex\hbox{E}\kern-.125emX}}
\begin{document}

\title{SmartPilot: A Multiagent CoPilot for Adaptive and Intelligent Manufacturing\\
}
\author{\IEEEauthorblockN{Chathurangi Shyalika}
\IEEEauthorblockA{\textit{Artificial Intelligence Institute} \\
\textit{University of South Carolina}\\
Columbia, SC 29208, USA \\
jayakodc@email.sc.edu}
\and
\IEEEauthorblockN{Renjith Prasad}
\IEEEauthorblockA{\textit{Artificial Intelligence Institute} \\
\textit{University of South Carolina}\\
Columbia, SC 29208, USA \\
kaipplir@mailbox.sc.edu}
\and
\IEEEauthorblockN{Alaa Al Ghazo}
\IEEEauthorblockA{\textit{Artificial Intelligence Institute} \\
\textit{University of South Carolina}\\
Columbia, SC 29208, USA \\
alaaalghazo@gmail.com}
\and
\IEEEauthorblockN{Darssan Eswaramoorthi}
\IEEEauthorblockA{\textit{Artificial Intelligence Institute} \\
\textit{University of South Carolina}\\
Columbia, SC 29208, USA \\
darssan@email.sc.edu}
\and
\IEEEauthorblockN{Harleen Kaur}
\IEEEauthorblockA{\textit{Artificial Intelligence Institute} \\
\textit{University of South Carolina}\\
Columbia, SC 29208, USA \\
harrs.bagga8@gmail.com}
\and
\IEEEauthorblockN{Sara Shree Muthuselvam}
\IEEEauthorblockA{\textit{Artificial Intelligence Institute} \\
\textit{University of South Carolina}\\
Columbia, SC 29208, USA \\
muthuses@email.sc.edu}
\and
\IEEEauthorblockN{Amit Sheth}
\IEEEauthorblockA{\textit{Artificial Intelligence Institute} \\
\textit{University of South Carolina}\\
Columbia, SC 29208, USA \\
amit@sc.edu}
}

\maketitle

\begin{abstract}

In the dynamic landscape of Industry 4.0, achieving efficiency, precision, and adaptability is essential to optimize manufacturing operations. 
Industries suffer due to supply chain disruptions caused by anomalies, which are being detected by current AI models but leaving domain experts uncertain without deeper insights into these anomalies. Additionally, operational inefficiencies persist due to inaccurate production forecasts and the limited effectiveness of traditional AI models for processing complex sensor data. Despite these advancements, existing systems lack the seamless integration of these capabilities needed to create a truly unified solution for enhancing production and decision-making. We propose SmartPilot, a neurosymbolic, multiagent CoPilot designed for advanced reasoning and contextual decision-making to address these challenges. SmartPilot processes multimodal sensor data and is compact to deploy on edge devices. It focuses on three key tasks: anomaly prediction,
production forecasting, and domain-specific question answering. By bridging the gap between AI capabilities and real-world industrial needs, SmartPilot empowers industries with intelligent decision-making and drives transformative innovation in manufacturing. The demonstration video, datasets, and supplementary materials are available at 
\url{https://github.com/ChathurangiShyalika/SmartPilot}
\end{abstract}


\vspace{-2 mm}
\section{Introduction}
Manufacturing processes in the Industry 4.0 era rely heavily on data-driven technologies for decision-making. However, challenges such as predictive analytics, supply chain disruptions, and inventory discrepancies hinder operational efficiency \cite{singh2024fundamental, nieberl2024review}. While foundational models (FMs) such as large language models (LLMs) have demonstrated success in text and image domains, their generalization to sensor data is restricted by challenges like heterogeneous datasets, data leakage, and the scarcity of labeled data. Hence, time-series-focused FMs like Time-MOE \cite{shi2024time} and MOIRAI \cite{gerald2024unified} often fail to generalize across manufacturing-specific tasks, hindering their practical utility \cite{shyalika2024time}. The rise of smaller and more efficient models \cite{hu2024minicpm} that can be locally or edge device deployed, offer lower cost, privacy and security chacheristic. When used to support a CoPilot, we can empower advanced manufacturing  applications. 

A CoPilot is an intelligent assistant that collaborates with users by providing context-aware, task-specific support and actionable insights \cite{woods2024if, sarkar2024copilot, roy2024neurosymbolic}. It can address complex enterprise challenges beyond the capabilities of basic conversational agents. SmartPilot is a CoPilot that uses a neurosymbolic and multiagent architecture designed to address industry-specific challenges. We address three critical manufacturing challenges associated with real-time operations: anomaly prediction, production forecasting, and domain-specific question answering. The system incorporates three specialized agents, each dedicated to one of these key functions. \textit{PredictX} agent is responsible for anomaly prediction and focuses on identifying irregularities in operational data to prevent disruption. \textit{Foresight} agent forecasts next-hour production and anticipates future production requirements to optimize resources. \textit{InfoGuide} agent 
facilitates real-time information retrieval tailored to manufacturing-specific questions for manufacturing professionals.

All three agents within SmartPilot are designed with lightweight models optimized for deployment on edge devices. This design ensures individual agents can perform real-time operations effectively in resource-constrained environments. The system integrates neurosymbolic techniques, combining neural network-based statistical methods with symbolic knowledge-based approaches \cite{Sheth2023NeurosymbolicA}. 
This work uses a neurosymbolic AI-based Knowledge-Infused Learning (KIL)\cite{sheth2019shades} for anomaly prediction and production forecasting.
Knowledge-infusion allows to integrate domain knowledge
with data-driven approaches, providing a foundation for more
robust and interpretable AI models \cite{sheth2019shades}. We use multimodal data as input for statistical methods and manufacturing ontologies as the foundation for symbolic reasoning. This gives SmartPilot the ability to give contextually aware and precise decision-making with high accuracy. SmartPilot is currently deployed in two manufacturing environments: a rocket assembly facility (in a full-function academic setup) and a vegemite production line (in a commercial production environment).

The main contributions of this paper are as follows:
\vspace{-1 mm}
\begin{itemize}[leftmargin=3mm]  % Adjusts left margin
    \item \textbf{Development of SmartPilot:} We introduce SmartPilot, a CoPilot specifically designed to address the unique challenges in manufacturing.
    
    \item \textbf{Innovative algorithms:} We present novel algorithms for anomaly prediction, production forecasting, and domain-specific question answering tailored to manufacturing needs.
    
    \item \textbf{Multiagent system design:} SmartPilot operates as a multiagent system, providing modular functionality and extensibility to address additional manufacturing-specific challenges.

    \item \textbf{Neurosymbolic methods:} 
    We use a neurosymbolic AI method involving domain or application-specific knowledge infusion for anomaly prediction and production forecasting.

    \item \textbf{Enhanced explainability:} Advanced techniques for user-level explainability are integrated into SmartPilot, including an ontology-based method, to ensure interpretable and trustworthy predictions.
    
    \item \textbf{Comprehensive evaluation:} We employ a robust evaluation framework comprising both numerical metrics and a user study. Quantitative assessments and qualitative feedback collectively validate the effectiveness and efficiency of SmartPilot.
\end{itemize}

\vspace{-4 mm}
\section{Related Work}
\subsection{LLMs in Manufacturing}
LLMs have emerged as transformative tools in manufacturing, offering the potential for innovative solutions in various applications. These models can improve tasks such as predictive maintenance, quality control, supply chain management, and process automation. LLMs like GPT-4V demonstrate capabilities in processing large-scale, heterogeneous manufacturing data, enabling real-time insights and operational efficiencies \cite{li2024large, kurkute2023scalable}. Their integration with advanced techniques such as federated and transfer learning addresses challenges such as data heterogeneity and legacy constraints of the system. In robotics, LLMs facilitate autonomous decision-making and task execution by integrating embodied intelligence for applications such as tool path design and industrial simulations\cite{fan2024embodied}. In addition, LLMs improve supply chain resilience through demand forecasting, logistics automation, and supplier risk management, leveraging their capacity to process unstructured data and ensure compliance with industry standards\cite{krishnamoorthy2024integrating}. Despite their potential, implementing and deploying LLMs in manufacturing requires addressing the need for highly specialized and intricate data, scalability, model adaptability, and real-time processing challenges \cite{li2024large,xia2024leveraging,shyalika2024time} 

\vspace{-2 mm}
\subsection{Agent-based Systems in Manufacturing}
Agent-based systems are transforming manufacturing by improving flexibility, coordination, and adaptability in dynamic environments. Leveraging LLMs, multi-agent systems can process context-specific instructions and enhance decision-making through natural language integration. For example, LLM-using frameworks allow for precise execution of tasks, such as G-code allocation, while improving agent communication and operational adaptability \cite{lim2024large}. Dynamic decision-making for resource reallocation is another key application. Multi-agent architectures with risk assessment capabilities enable real-time rescheduling to handle disruptions such as machine breakdowns, balancing flexibility, and computational efficiency \cite{bi2024dynamic}. These systems also optimize manufacturing processes by integrating resource agents for data-driven decision-making, improving efficiency, reducing costs, and improving service quality \cite{liu2024manufacturing}. Furthermore, AI-enabled agents in manufacturing of advanced materials demonstrate their potential in smart design and defect detection, advancing intelligent manufacturing \cite{wang2024artificial}. Lastly, agent-based collaboration empowered by digital platforms enhances resilience and linkage across manufacturing enterprises, driven by dynamic incentive mechanisms\cite{yi2024manufacturing}.

\vspace{-2 mm}
\subsection{AI-driven Copilots in Manufacturing}
AI-driven copilots, powered by the advancements in generative AI, can transform manufacturing by improving decision making, safety, and operational efficiency. Generative AI copilots support safety experts in human-robot collaborations by identifying hazards, their causes, and proposing mitigation strategies. This integration streamlines hazard analysis and improves safety protocols \cite{kranz2024generative}. In automation equipment selection, copilots leverage LLMs with Retrieval-Augmented Generation (RAG) to provide structured and traceable decision-making processes \cite{werheid2024designing}. For business optimization, AI copilots fine-tuned with modular prompt engineering synthesize complex problem formulations in production scheduling \cite{amarasinghe2023ai}. This approach minimizes reliance on human expertise while maintaining high accuracy and adaptability. AI chatbots further enhance operations, supply chain management, and logistics by improving efficiency, responsiveness, and intelligence in decision-making, though workforce adaptation remains a challenge \cite{durach2024hello}. These AI-driven copilots demonstrate the potential for intelligent, responsive, and efficient manufacturing systems, paving the way for more adaptive and innovative operational paradigms.
%\subsection{Anomaly prediction}
%\subsection{Production forecasting}
%\textcolor{red}{Prof.Alaa. please write this section in detail}
%\subsection{QA retrieval}

%single system that performs each of the above function unavailable in the industry.
%need of a such system-copilot

Existing approaches focusing on LLMs, agents, and AI-driven copilots in manufacturing are highly specialized, addressing narrow domains. 
Existing systems limit their applicability to broader manufacturing challenges, highlighting a significant research gap due to the lack of a unified system capable of addressing multiple manufacturing-specific use cases within a single framework. In addition, these systems lack integration, interpretability, scalability, and adaptability to diverse and complex scenarios, further constraining their utility. Developing such an integrated AI-driven copilot is essential to streamline operations, enhance scalability, and meet the dynamic demands of modern manufacturing environments.

\vspace{-2mm}
\section{Core System Framework} 
The SmartPilot framework is a neurosymbolic, agent-based system designed to integrate multimodal data for real-time manufacturing process optimization. It offers a unified and extensible solution to address a broader range of manufacturing-specific challenges. This section describes the key components and technical features that define the SmartPilot framework.

\subsection{Multiagent Architecture} 
\begin{figure}[!htb]
  \centering
\includegraphics[width=0.99\linewidth]{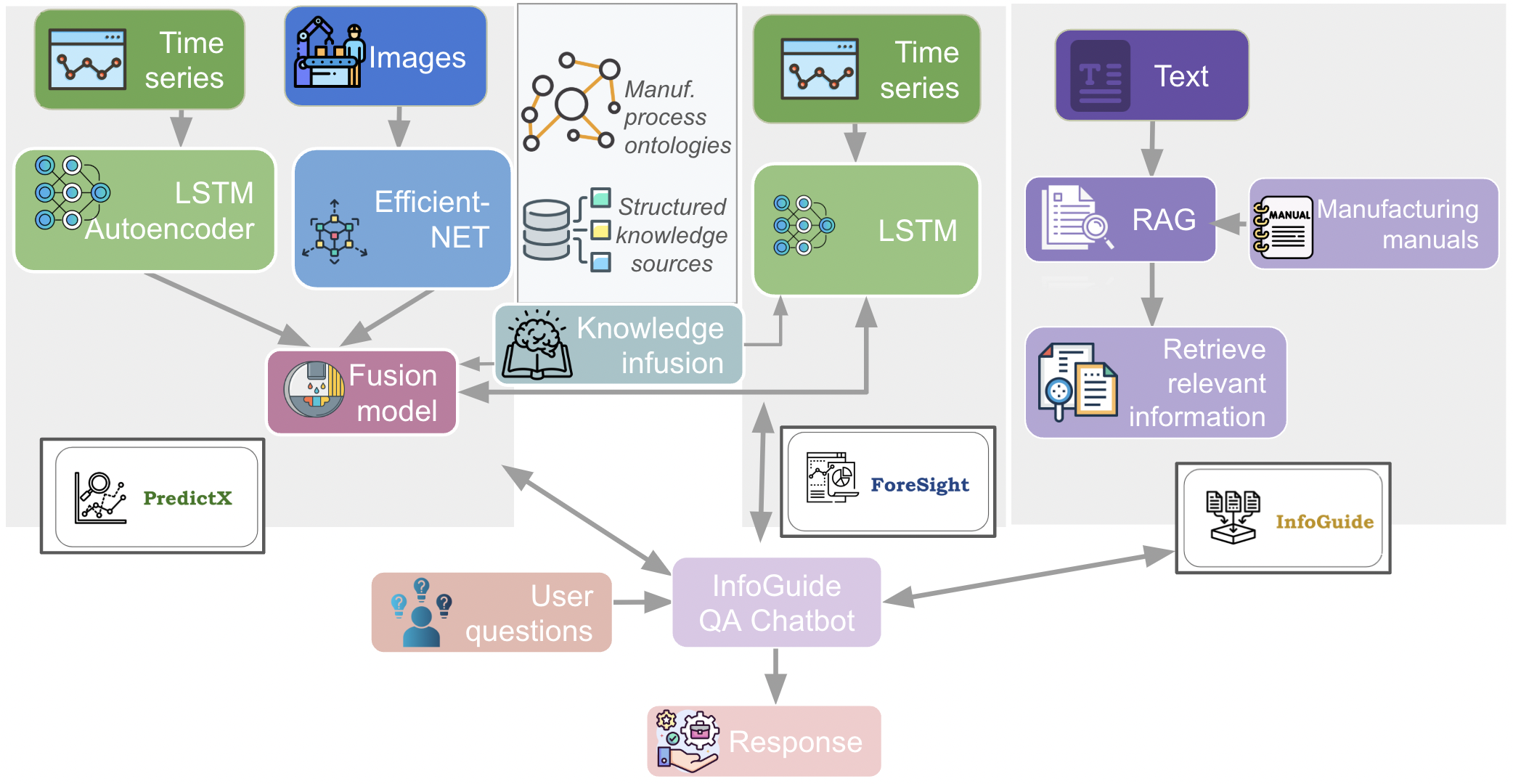}
\caption{Agent-based architecture of SmartPilot and their interactions}
  \label{fig:overall_architecture1}
\end{figure}

SmartPilot’s architecture is built around three specialized agents and a structured framework for integration, as shown in Figure \ref{fig:overall_architecture1}. 
%These components collectively address critical challenges in manufacturing.

\subsubsection{PredictX: Anomaly Prediction Agent} 
\begin{figure*}[!htb]
  \centering
  \vspace{-4 mm}\includegraphics[width=0.999\linewidth]{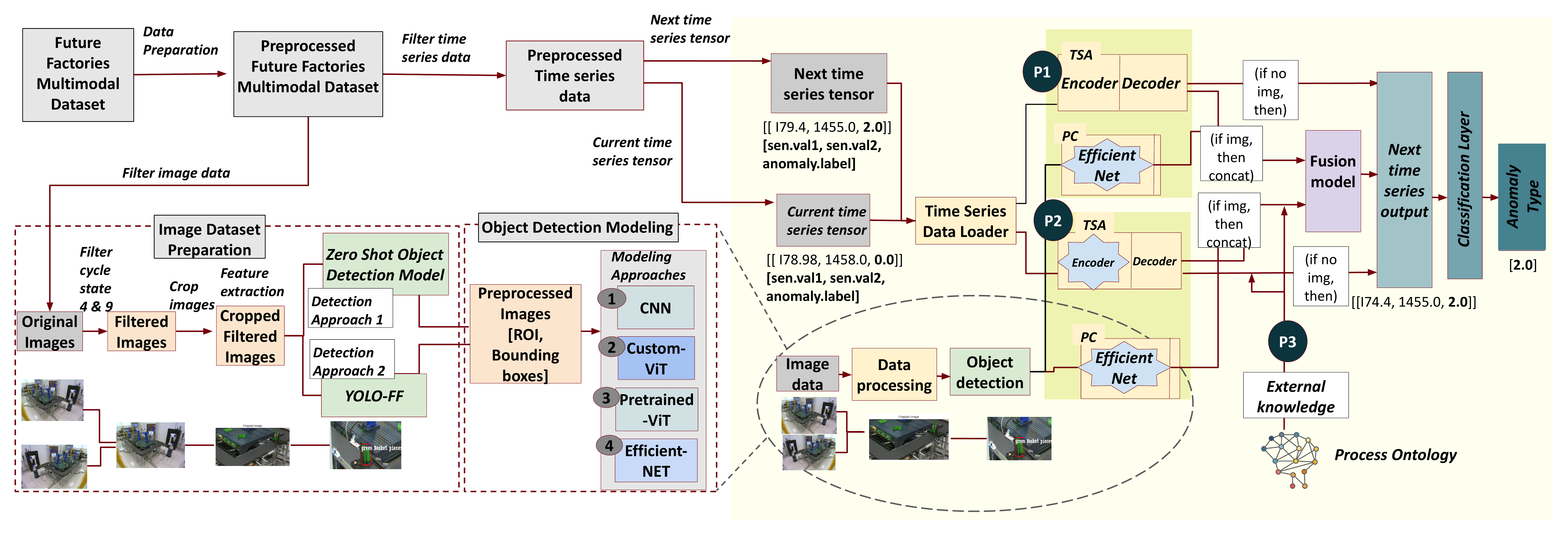}
  \vspace{-4 mm}
\caption{
Architecture of PredictX agent:
It integrates time series data and image inputs for anomaly prediction through a multi-stage process. The system begins with preprocessing and feature extraction, utilizing a pretrained EfficientNet (PC) model for image features and a time series autoencoder (TSA) for time series data. The extracted features are then fused, incorporating external process ontology knowledge to enhance the model's predictive capabilities. The fusion model ultimately predicts the next time series output and classifies the anomaly types. Three baseline approaches are implemented for comparison: P1, a Decision-Level Fusion approach; P2, a Decision-Level Fusion with Transfer Learning (where the hyperparameters of the autoencoder and EfficientNet-B0, training process, and loss function remain consistent with P1, but the encoder is frozen to prevent gradient updates); and P3, an Enhanced Decision-Level Fusion with Transfer Learning via Neurosymbolic AI. In P3, a custom loss function is introduced that combines Weighted Mean Squared Error (WMSE) loss with an additional penalty, infusing external knowledge on sensor ranges derived from the process ontology.}
  \label{fig:overall_architecture}
  \vspace{-4 mm}
\end{figure*}

In PredictX \footnote{https://github.com/smtmnfg/NSF-MAP}, we introduce a fusion-based approach for multimodal anomaly prediction in assembly pipelines, as shown in Figure \ref{fig:overall_architecture}. Specifically, we use decision-level fusion (a.k.a. late fusion), a fusion technique that combines the decisions of multiple classifiers into a shared decision \cite{roggen2013signal}. The fusion model integrates time series data, processed via an autoencoder, with image data analyzed using a fine-tuned EfficientNet and combines the outputs of these models through a fully connected network.  
We utilize a transfer learning approach, where a pre-trained model is adapted for a new task \cite{torrey2010transfer, pan2009survey}. In this approach, the encoder of the autoencoder is frozen, and only the decoder is trained. This strategy helps mitigate the limitations of training the entire autoencoder-decoder network end-to-end, which may lead to overfitting, poor generalization to unseen data, longer training times, and increased computational costs. As the neurosymbolic approach, we use manufacturing-based process ontologies to infuse domain-specific knowledge in the model training process and to provide user-level explanations for the predictions. The PredictX model is currently trained on multimodal data from the Future Factories (FF) dataset, which includes data from a rocket assembly process \cite{harik2024analog, Harik_2024}. Section \ref{sec:app} discusses its detailed application use case.

\subsubsection{ForeSight: Production Forecasting Agent} 
\begin{figure}[!htb]
  \centering
  \includegraphics[width=0.89\linewidth]{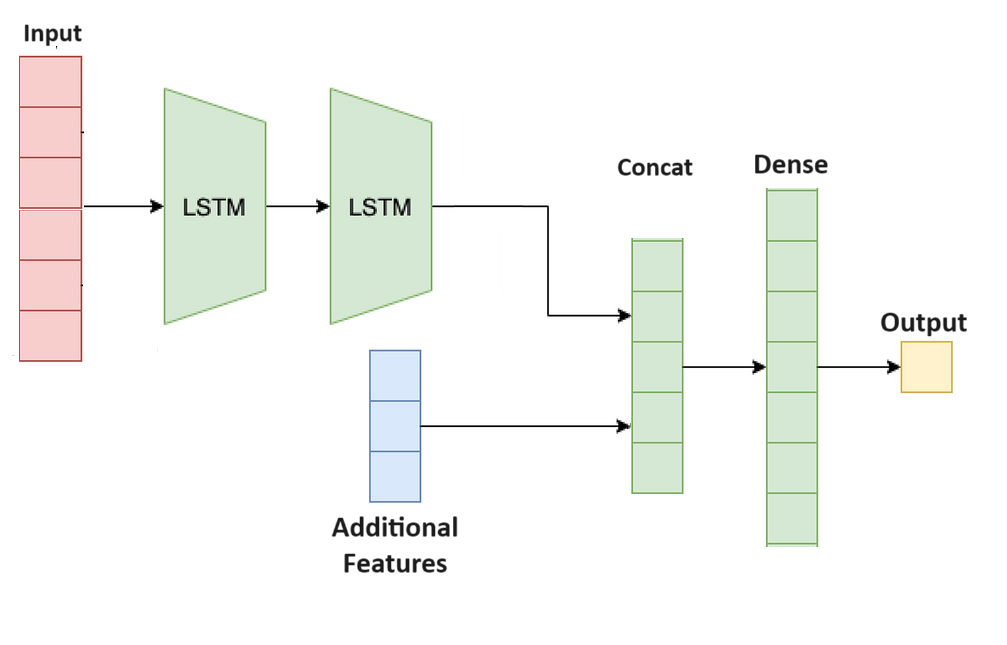}
  \vspace{-4 mm}
\caption{Architecture of ForeSight agent: 
It utilizes an LSTM model for production forecasting, using historical data from the target variables. The architecture includes two LSTM layers to capture temporal dependencies, with additional features infused at the dense layer level.}
  \label{fig:overall_architecture_a2}
\end{figure}

ForeSight (Figure \ref{fig:overall_architecture_a2})  utilizes a Long Short-Term Memory (LSTM)-based model for production forecasting, designed to capture temporal dependencies while integrating domain-specific knowledge for enhanced accuracy. The model processes historical data from target variables—such as production metrics—and supplements this with structured features like raw material quantities, process-based ratios, and other contextual metrics specific to the production process. This combination enables the model to learn both temporal patterns and process-specific relationships. The architecture includes two LSTM layers, where the first outputs sequences and the second produces a consolidated vector capturing temporal dependencies. This output is then concatenated with the structured features, providing a rich, context-infused representation. The resulting vector is processed through a dense layer  leading to the final output layer that predicts the target variables. 
The ForeSight agent is currently trained on the vegemite evaporation dataset \cite{banerjee2021iiot, banerjee2024improving} and analog data derived from the FF dataset \cite{harik2024analog}. Detailed application use cases are provided in Section \ref{sec:app}.

\subsubsection{InfoGuide: Domain-Specific Q\&A Agent} 

\begin{figure}[!htb]
  \centering
  \vspace{-4 mm}\includegraphics[width=0.999\linewidth]{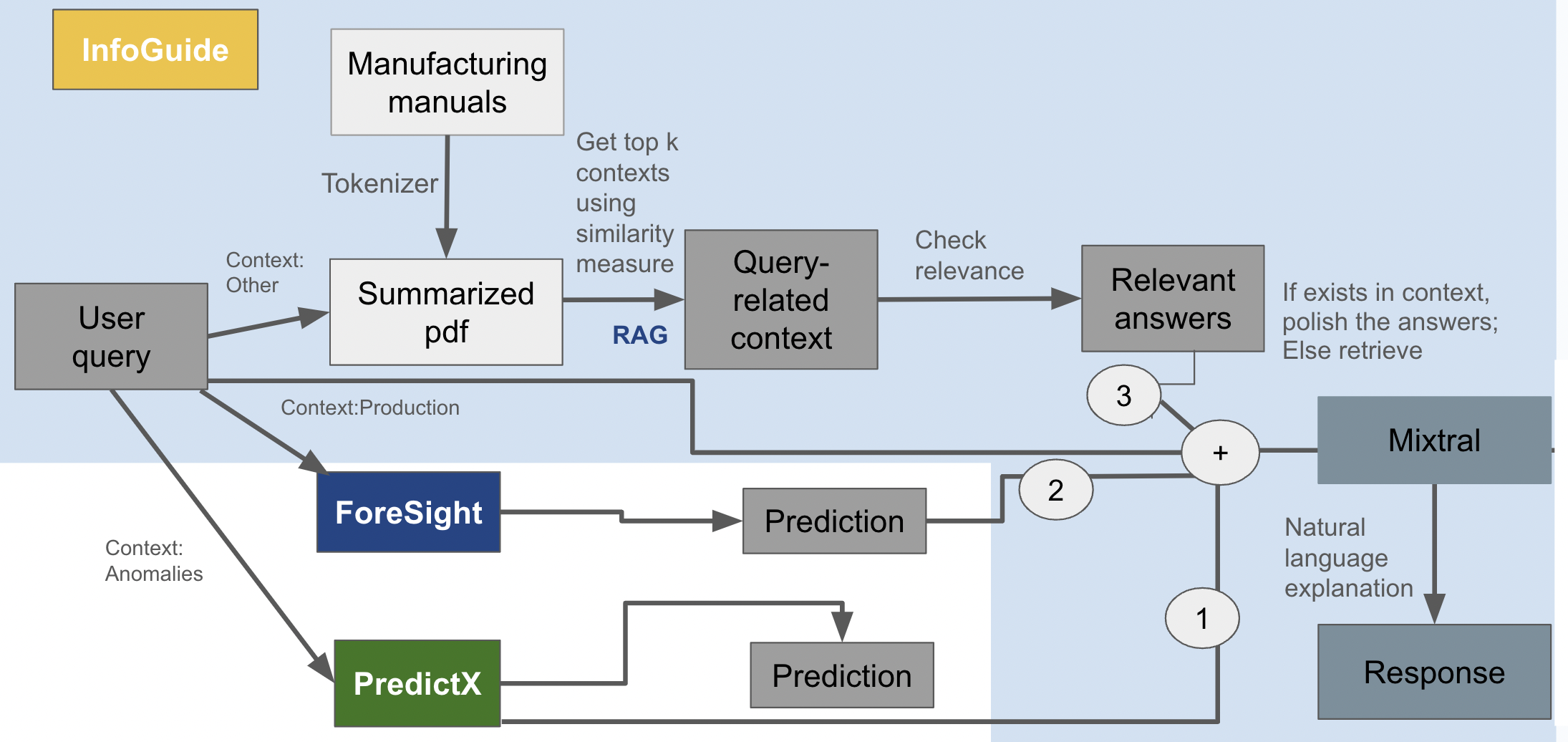}
\caption{Architecture of InfoGuide agent: The blue area indicates the boundary of the InfoGuide agent. InfoGuide agent also interacts with the PredictX and ForeSight agents to respond to queries related to real-time anomaly prediction and production forecasting.}
  \label{fig:overall_architecture_a3}
  \vspace{-6 mm}
\end{figure}

InfoGuide leverages meaningful texts derived from manufacturing manuals to provide answers to contextually rich manufacturing-specific questions.
Figure \ref{fig:overall_architecture_a3} presents the overall architecture of the InfoGuide agent. To summarize the manufacturing manuals in PDF for efficient retrieval, we first converted the PDF to text using \textit{pdfplumber}, which effectively removes images and tables. Next, we cleaned the text by removing noise such as page numbers, headers, footers, titles, and irrelevant symbols using regex. We then employed \textit{CharacterTextSplitter} to divide the text into smaller chunks based on paragraph splits. Identifying relevant keywords like "safety," "maintenance," "operation," "installation," "inspection," "warning," "danger," and "caution," we expanded this list with lemmatized synonyms. Using cosine similarity, we retrieved chunks corresponding to these keywords, setting a threshold based on experimental results. Finally, we used the \textit{LLaMA-2-7b-chat} model to summarize these chunks, which were then used as the context for further retrieval. This context is referred to the \textit{Summarized PDF} in Figure \ref{fig:overall_architecture_a3}.

When a user submits a query, the system uses RAG to find the top k relevant contexts based on similarity measures: Neural context retrieval (using cosine similarity with BERT embeddings) and Symbolic context retrieval (using Jaccard similarity with tokenized keywords). If the similarity score exceeds a set threshold, the user query, agent-specific template and retrieved context are sent to the Mixtral language model to generate a coherent response. If the score falls below the threshold, the system refrains from providing an answer, effectively mitigating the risk of generating hallucinated responses.

\subsubsection{Inter-Agent Connectivity}
These three agents are seamlessly connected, fostering a cohesive system for anomaly prediction, production forecasting, and information retrieval.
PredictX feeds anomaly-related insights into ForeSight, allowing the forecasting model to adapt to the evolving production system. This interconnected approach enables real-time dynamic responses to user queries, ensuring that both anomaly prediction and production forecasting are aligned with changing operational conditions. The model is further integrated with InfoGuide for effective information retrieval, enhancing the system's ability to provide actionable insights.
To handle the interconnection between the InfoGuide and the other two agents, we fine-tune the DistilBERT language model on the outputs of PredictX and Foresight, employing Low-Rank Adaptation (LoRA). The fine-tuned model is integrated with InfoGuide, enabling effective responses to user queries about anomaly prediction and production forecasting through real-time information retrieval.

\vspace{-2mm}
\subsection{Multimodal Data Integration} SmartPilot processes diverse data types, including time series sensor readings, image data, and text. Multimodal data is used in the autoencoder-based fusion model in PredictX (time series data and image data) and the data flow architecture of InfoGuide (time series data, image data, and text data) agents,  which combine and interpret heterogeneous data streams. This approach ensures comprehensive insights for anomaly prediction, production forecasting, and information retrieval.

\subsection{Custom, Compact, and Neurosymbolic AI Model} SmartPilot's AI model is tailored for manufacturing applications, emphasizing compactness and neurosymbolic integration. The custom aspect of the model involves the optimization of architectures tailored to specific domain tasks, utilizing curated datasets and manufacturing ontologies. Its compact nature ensures the use of lightweight architectures, reducing computational demands and facilitating deployment on edge devices with minimal resource consumption. Additionally, the model incorporates neurosymbolic techniques, blending statistical methods with symbolic reasoning to improve both the accuracy of decision-making processes and the interpretability of the model's outcomes. 

\section{Applications and Case Studies} 
\label{sec:app}

\subsection{Deployment Setup of SmartPilot}
\begin{figure}[!htb]
  \centering
\includegraphics[width=0.99\linewidth]{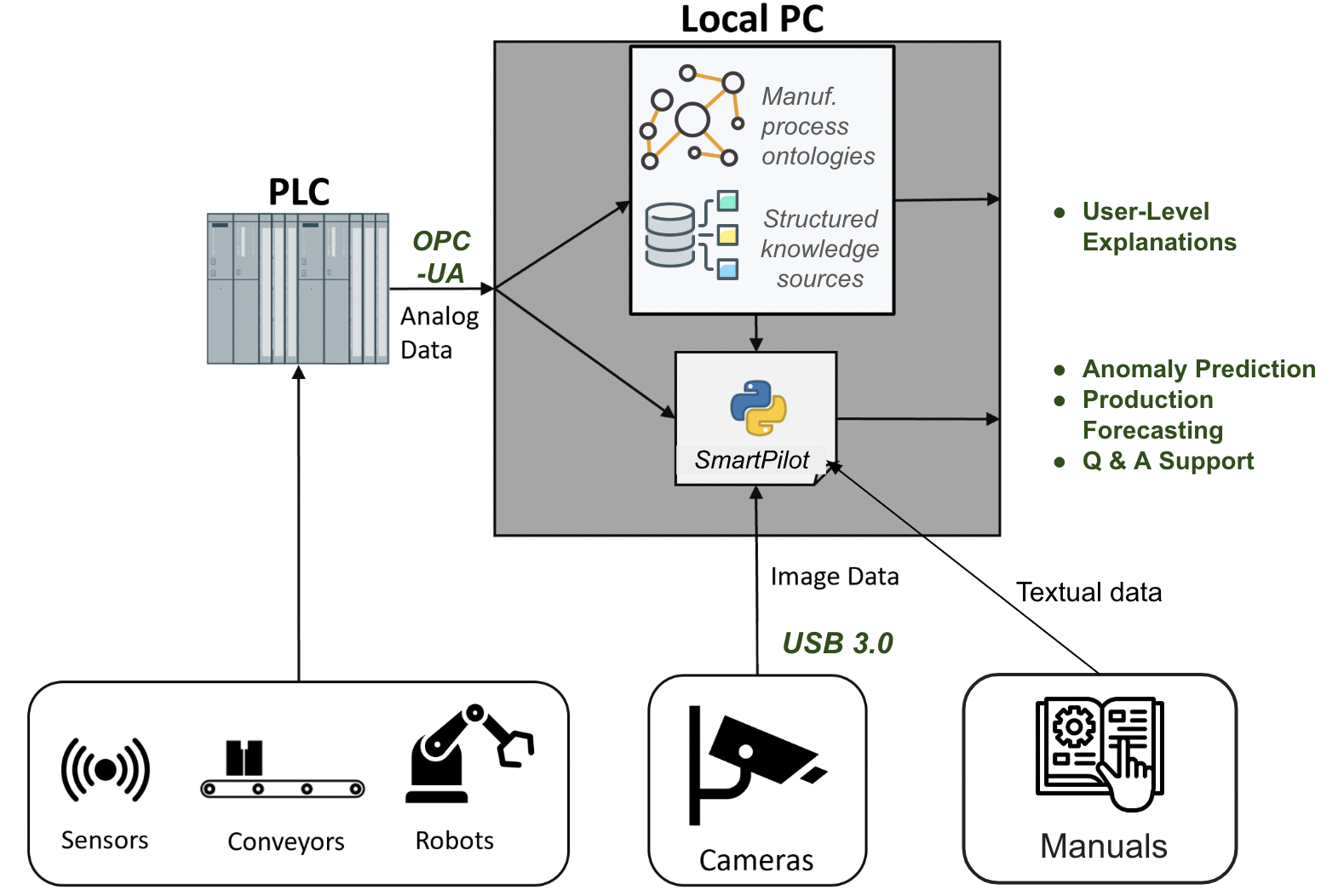}
\caption{Deployment Setup of SmartPilot}
  \label{fig:dep}
  \vspace{-4 mm}
\end{figure}

Figure \ref{fig:dep} illustrates the architecture for real-time deployment of the SmartPilot. It highlights integrating the SmartPilot system that includes the trained models with the Open Platform Communications Unified Architecture (OPC-UA) server for sensor data retrieval and the connection to cameras and manuals for image and textual data acquisition, enabling seamless real-time predictions. SmartPilot's capabilities have been deployed and demonstrated in two distinct manufacturing environments: toy rocket assembly and vegemite production, as outlined in the following sections. The
inference code, user interface code for deployment, and the demo of deployment are included in the supplementary files.
\vspace{-4 mm}

\subsection{Rocket Assembly Use Case} In the rocket assembly process \cite{harik2024analog, Harik_2024}, PredictX predicts anomalies caused by missing components in assembling a rocket, while ForeSight anticipates production based on assembly schedules. InfoGuide assists operators by answering technical questions about the assembly process, ensuring operational continuity.

The FF dataset consists of measurements of this rocket assembly pipeline, designed to meet industrial standards in deploying actuators, control mechanisms, and transducers. Its multimodal version includes synchronized images captured from two cameras positioned on opposite sides of the testbed, which continuously record the operations. We use the Dynamic Process Ontology\footnote{https://github.com/revathyramanan/Dynamic-Process-Ontology} designed and developed for this rocket assembly use case to support knowledge-infused learning and provide user-level explanations for the predictions made by the model. The domain-specific knowledge on sensors, cycle states, robots, and machinery are being infused at the model training stage of PredictX.  We also use Dynamic Process Ontology to provide user-level explanations for the
outputs generated by the model. In this case, a predicted anomaly can be explained as follows: (i) which variable is responsible for the anomaly? (ii) what were the function of the robots during that state? (iii) what are the
expected values of that variable in that state? The ontology can
also capture misclassification in certain scenarios. PredictX enables proactive identification of potential disruptions through these capabilities, reducing downtime and production losses. In addition, we use the manuals of the robots, sensors, conveyor belts, and the stands currently available in the facility. These will serve as inputs to the InfoGuide agent.

\vspace{-2 mm}
\subsection{Vegemite Production Use Case} In the Vegemite production line, ForeSight predicts production for the next hour using the assembly schedules of vegemite production. InfoGuide agent supports operators by answering technical questions about the production process, ensuring smooth operations.

The Vegemite Evaporation dataset \cite{banerjee2021iiot,banerjee2024improving} comprises detailed measurements collected from the industrial yeast processing pipeline used in the production of Vegemite, adhering to food-grade manufacturing standards. It captures a comprehensive range of parameters from industrial evaporators, including raw yeast input quantities, evaporation ratios (IS/TS), flow rates, temperature, and pressure settings. These measurements provide a granular view of the evaporation process, a critical stage in concentrating raw yeast into the final paste.
The dataset also includes product quality measurements such as the percentage of total solids in the paste, recorded at regular intervals using specialized refractometers. These measurements serve as precise indicators of the quality and consistency of the paste throughout the evaporation process. Alongside sensor data, the dataset includes batch identifiers, timestamps, and metadata on operational configurations, ensuring traceability and alignment with production records.
A unique feature of the dataset is its inclusion of variability in raw yeast types, reflecting seasonal and source-based differences in input materials. 
%This variability ensures the dataset's relevance to real-world manufacturing scenarios, capturing the inherent fluctuations in raw material properties. 
Additionally, the dataset contains calibration logs and machine state transitions, offering a complete view of the operational dynamics.

%The Vegemite Evaporation dataset, used to train the model, captures high-resolution data on the yeast processing stages involved in Vegemite production. 
The dataset includes key target variables (\textit{Yeast - BRD}, \textit{Yeast - BRN}, \textit{Yeast - FMX}) and supplementary features like raw yeast input quantities, evaporation ratios (IS/TS), and control settings for industrial evaporators. The dataset records product quality metrics such as the percentage of total solids, a critical indicator of product consistency. Variability in the dataset arises from factors like seasonal differences in yeast properties and operational adjustments made during production. This rich and dynamic dataset is preprocessed using normalization techniques to ensure compatibility with the LSTM framework, enabling ForeSight to accurately forecast production outcomes and adapt to the complex dynamics of the Vegemite manufacturing process. Additionally, we use manuals for the machines, sensors, conveyor belts, stands available in the facility, and supplementary resources on the yeast evaporation process. These inputs will be utilized by the InfoGuide agent.

\vspace{-1 mm}
\subsection{Demonstration Details of SmartPilot}
\vspace{-1 mm}
This section provides a detailed demonstration of SmartPilot, focusing on its user interfaces, real-time integration with backend systems, and the retrieval of user-level explanations for predictions. The user interfaces were developed using HTML, CSS, JavaScript and Streamlit.

\subsubsection{User Interfaces and Interactivity} 
\begin{figure}[!htb]
  \centering
  \vspace{-4 mm}\includegraphics[width=0.89\linewidth]{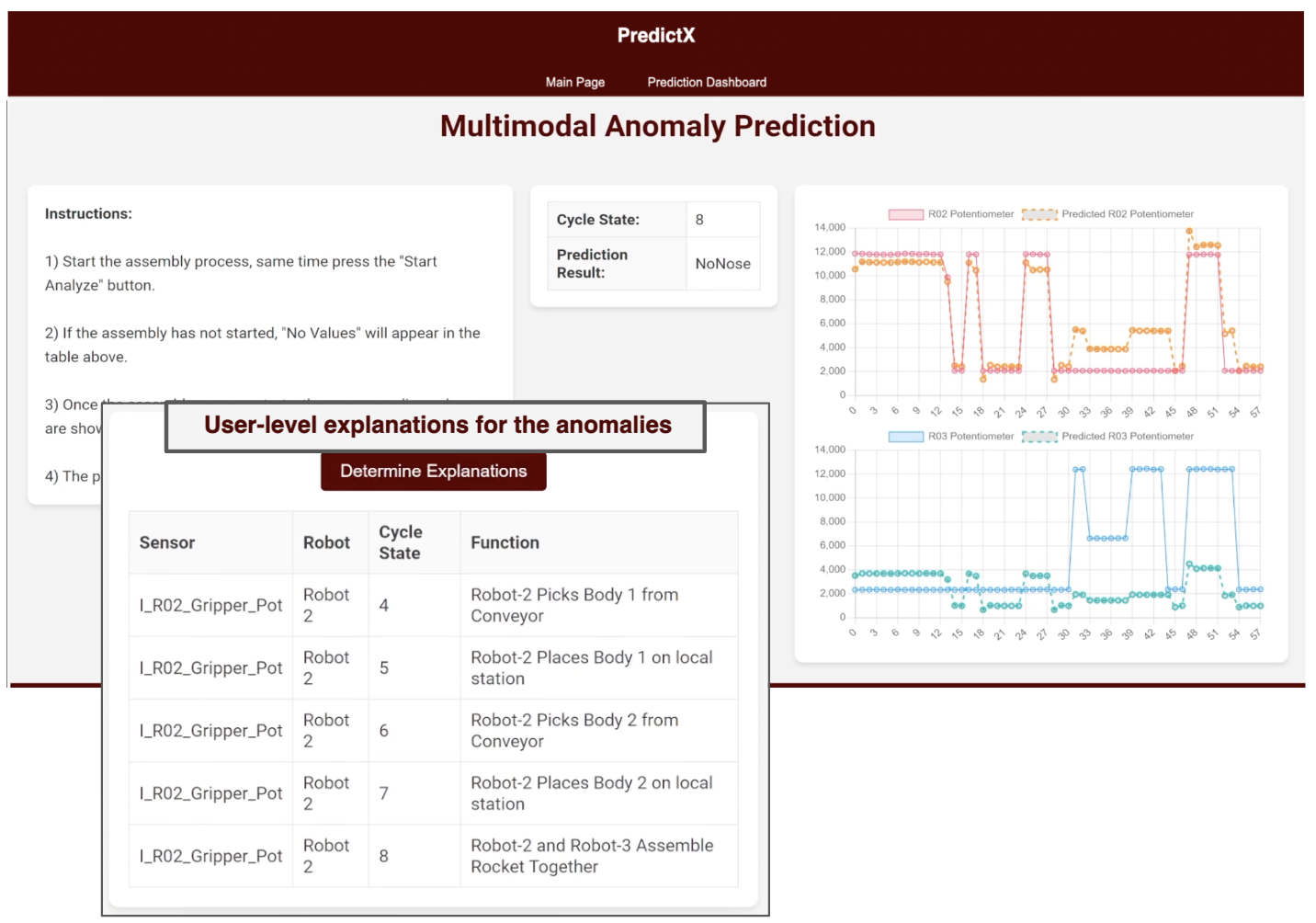}
  \vspace{-4 mm}
\caption{Real-time anomaly prediction and user-level explanations given by PredictX agent, with a similar design applied to ForeSight agent}
  \label{fig:predictx}

\end{figure}

\begin{figure}[!htb]
  \centering
  \vspace{-4 mm}\includegraphics[width=0.89\linewidth]{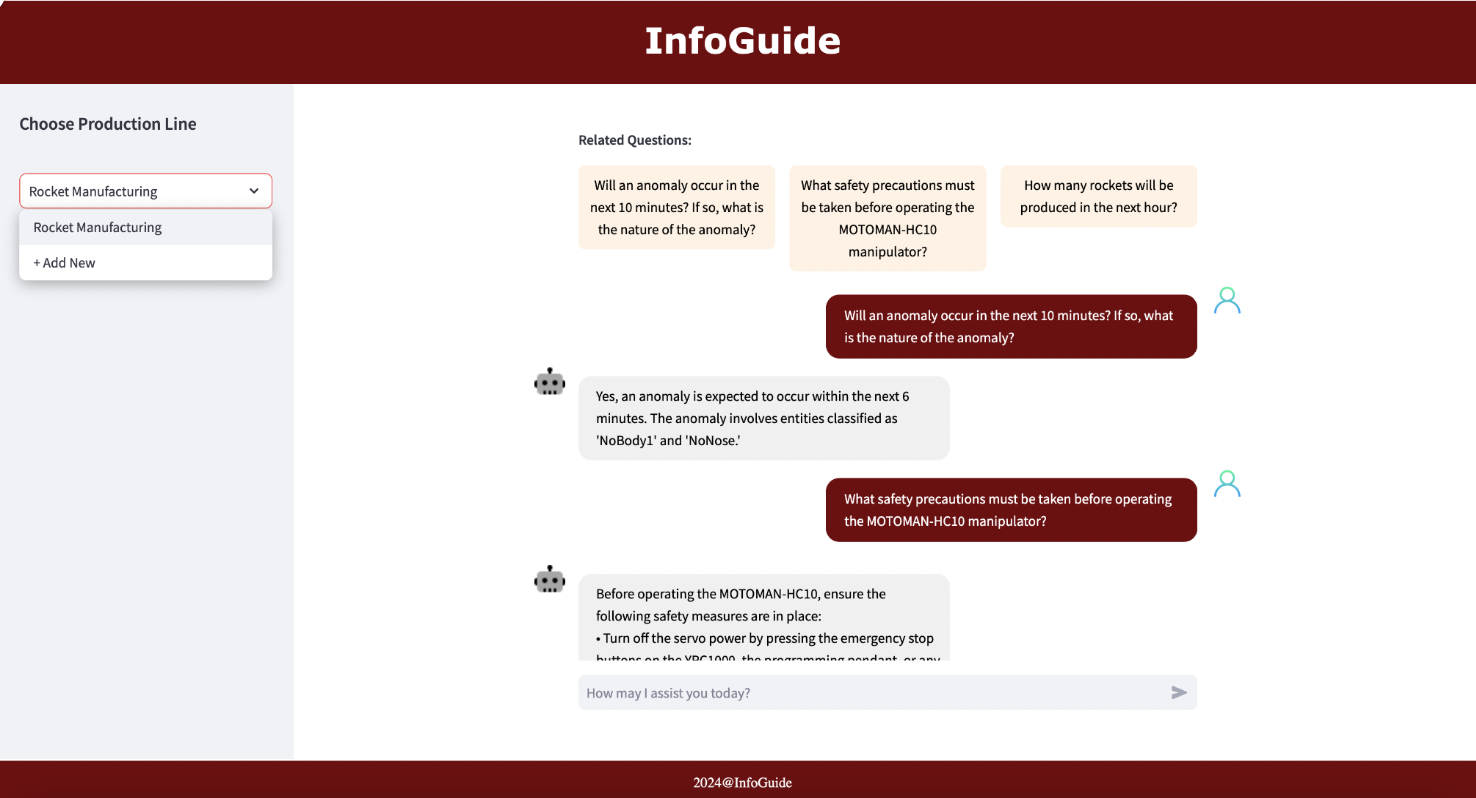}
  \vspace{-2 mm}
\caption{User interface of InfoGuide: Providing real-time responses to user queries, with an option to add a new facility from the left panel}
  \label{fig:infoguide}
  \vspace{-4 mm}
\end{figure}

SmartPilot offers a responsive, modular, and user-friendly interface for efficient interaction. Upon accessing, the main interface provides three buttons to access the three agents. PredictX offers two options: the prediction dashboard for real-time anomaly prediction and the explanation dashboard for detailed insights into the predictions, as shown in Figure \ref{fig:predictx}. ForeSight interface features the forecasting dashboard, enabling users to monitor and forecast future production. InfoGuide interface serves as a real-time question-and-answer chatbot, allowing users to input domain-specific questions and receive immediate responses as shown in Figure \ref{fig:infoguide}. 

\subsubsection{Real-time Integration with Backend Systems}
The prediction and forecasting dashboards in PredictX and ForeSight enable them to connect with an OPC-UA server, allowing them to visualize real-time sensor data for anomaly prediction and production forecasting, respectively. InfoGuide is similarly connected to the OPC-UA server, utilizing real-time data to provide accurate responses to queries related to anomaly prediction and production forecasting.

\subsubsection{User-level Explanation of Predictions}
Once anomalies are predicted, the second window of PredictX provides user-level explanations, including details on (i) which variables are responsible for the anomaly, (ii) the functions performed by the robots during that state and (iii) the expected values of those variables. 

\section{Results and Evaluation} SmartPilot is evaluated on real-world datasets from the aforementioned use cases. Metrics such as prediction accuracy, downtime reduction, and user satisfaction were used to assess its performance. For the quantitative evaluations, both the preprocessed datasets are divided into 80\% for training and 20\% for testing. 

\subsection{Anomaly Prediction Results of PredictX} 
\begin{figure}[!htb]
  \centering
\includegraphics[width=0.99\linewidth]{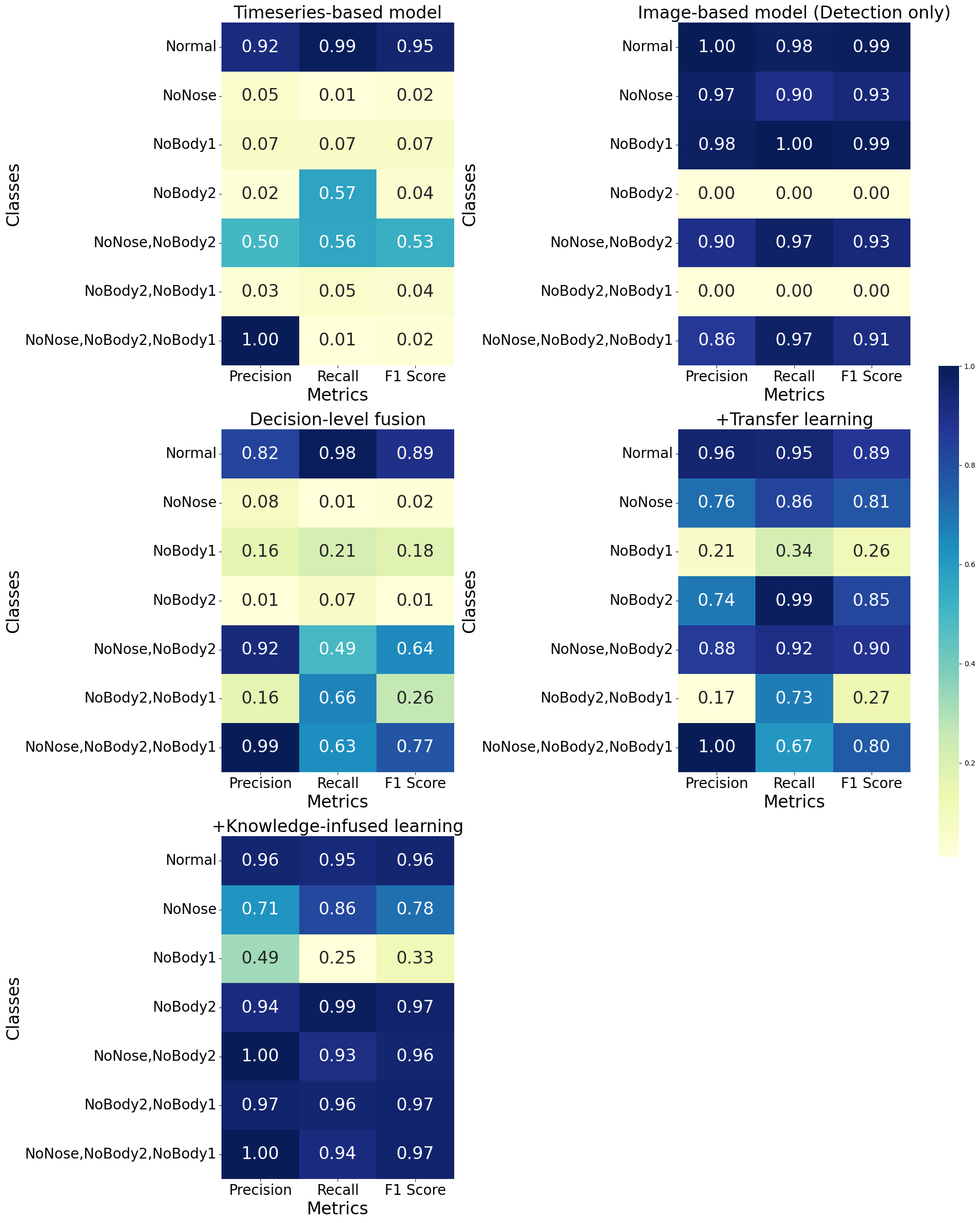}
  \vspace{-6 mm}
\caption{Experimental Results of Predicting Different Anomaly Types by PredictX. The FF dataset has six types of anomalies; [NoNose], [NoBody1], [NoBody2], [NoNose, NoBody2], [NoBody2, NoBody1], and [NoNose,NoBody2,NoBody1] respectively. [Normal] category represents instances without any anomalies.
}
  \label{fig:ANOMALY_TYPES}
  \vspace{-4 mm}
\end{figure}

PredictX achieved an accuracy of 93\% in predicting anomalies regarding to missing parts in rocket assembly use case, demonstrating its effectiveness. Table \ref{tab:results} summarizes the results of our experiments on the test set and the ablation studies across different baselines. 
We implement four baseline models for the PredictX agent: a time series-based Autoencoder model (B1), an image-based EfficientNet-B0 model (B2), a Decision-Level Fusion approach (P1), and Decision-Level Fusion with Transfer Learning (P2). The proposed approach, Enhanced Decision-Level Fusion with Transfer Learning through Neurosymbolic AI (DLF+TL+KIL), is denoted as P3 in Table \ref{tab:results}. We evaluate the performance using four metrics: weighted averages of precision, recall, F1-score, and accuracy. The weighted averages are calculated based on the six types of anomaly classes and the normal class. The Decision level fusion model achieves 72\% overall accuracy and 76.05\% of precision, 72\% of recall, 72.03\% of F1 score, respectively. With the inclusion of Transfer learning module, the accuracy and the F1 score increased by 39\% and 45\%, respectively. Similarly, infusion of knowledge on sensors enhanced the accuracy and the F1 score by 47\% and 52\%, respectively. Figure \ref{fig:ANOMALY_TYPES} depicts the performance of detecting various anomaly types and the normal class across various modeling approaches. It can be observed that among all the models, Decision level fusion with transfer learning and knowledge-infused learning approach gives the best results in detecting five out of seven types of classes.

\begin{table}
\centering
\scriptsize
\caption{Experimental Results of PredictX and Ablation Studies (mean(\%) ± std(\%)). Support: Number of samples used. Bold indicates the best performance.D: model used only for Detection.}
\label{tab:results}
\begin{tabular}{|c|c|c|c|c|c|} 
\hline
\textbf{Model}                                                                                                   & \begin{tabular}[c]{@{}c@{}}\textbf{Weighted }\\\textbf{Avg.}\\\textbf{Precision}\end{tabular} & \begin{tabular}[c]{@{}c@{}}\textbf{Weighted }\\\textbf{Avg.}\\\textbf{Recall}\end{tabular} & \begin{tabular}[c]{@{}c@{}}\textbf{Weighted }\\\textbf{Avg.}\\\textbf{F1-Score}\end{tabular} & \textbf{Accuracy}                                                                 & \textbf{*Support}  \\ 
\hline
B1                                                                                                               & \begin{tabular}[c]{@{}c@{}}74.00±\\1.00\%\end{tabular}                                        & \begin{tabular}[c]{@{}c@{}}63.00±\\1.00\%\end{tabular}                                     & \begin{tabular}[c]{@{}c@{}}61.00±\\1.00\%\end{tabular}                                       & \begin{tabular}[c]{@{}c@{}}63.00±\\1.00\%\end{tabular}                            & 33201              \\ 
\hline
B2: *D                                                                                                           & \begin{tabular}[c]{@{}c@{}}97.00±\\1.05\%\end{tabular}                                        & \begin{tabular}[c]{@{}c@{}}97.00±\\0.5\%\end{tabular}                                      & \begin{tabular}[c]{@{}c@{}}97.00±\\1.00\%\end{tabular}                                       & \begin{tabular}[c]{@{}c@{}}97.00±\\1.00\%\end{tabular}                            & 3119               \\ 
\hline
DLF (P1)                                                                                                         & \begin{tabular}[c]{@{}c@{}}76.05±\\2.00\%\end{tabular}                                        & \begin{tabular}[c]{@{}c@{}}72.00±\\1.70\%\end{tabular}                                     & \begin{tabular}[c]{@{}c@{}}72.03±\\2.05\%\end{tabular}                                       & \begin{tabular}[c]{@{}c@{}}72.00±\\2.00\%\end{tabular}                            & 33201              \\ 
\hline
\begin{tabular}[c]{@{}c@{}}DLF (TS+\\zero tensor\\image)\end{tabular}                                            & \begin{tabular}[c]{@{}c@{}}80.00±\\1.00\%\end{tabular}                                        & \begin{tabular}[c]{@{}c@{}}64.00±\\0.5\%\end{tabular}                                      & \begin{tabular}[c]{@{}c@{}}64.00±\\0.5\%\end{tabular}                                        & \begin{tabular}[c]{@{}c@{}}64.00±\\0.05\%\end{tabular}                            & 33201              \\ 
\hline
\begin{tabular}[c]{@{}c@{}}DLF+TL \\(P2)\end{tabular}                                                            & \begin{tabular}[c]{@{}c@{}}91.00±\\0.05\%\end{tabular}                                        & \begin{tabular}[c]{@{}c@{}}88.02±\\0.05\%\end{tabular}                                     & \begin{tabular}[c]{@{}c@{}}89.00±\\0.05\%\end{tabular}                                       & \begin{tabular}[c]{@{}c@{}}88.00±\\0.05\%\end{tabular}                            & 33201              \\ 
\hline
DLF+KIL                                                                                                          & \begin{tabular}[c]{@{}c@{}}93.00±\\1.00\%\end{tabular}                                        & \begin{tabular}[c]{@{}c@{}}90.00±\\0.05\%\end{tabular}                                     & \begin{tabular}[c]{@{}c@{}}91.00±\\0.05\%\end{tabular}                                       & \begin{tabular}[c]{@{}c@{}}90.00±\\0.05\%\end{tabular}                            & 33201              \\ 
\hline
\begin{tabular}[c]{@{}c@{}}\textbf{DLF+TL+}\\\textbf{KIL(P3)}\\\textbf{(PredictX }\\\textbf{model)}\end{tabular} & \begin{tabular}[c]{@{}c@{}}\textbf{94.00±}\\\textbf{1.00\%}\end{tabular}                      & \begin{tabular}[c]{@{}c@{}}\textbf{93.00±}\\\textbf{0.75\%}\end{tabular}                   & \begin{tabular}[c]{@{}c@{}}\textbf{93\textbf{.00}±}\\\textbf{0.05\%}\end{tabular}            & \begin{tabular}[c]{@{}c@{}}\textbf{93\textbf{.00}±}\\\textbf{1.00\%}\end{tabular} & \textbf{33201}     \\
\hline
\end{tabular}
\vspace{-4 mm}
\end{table}

\subsection{Production Forecasting Results of ForeSight} 

Evaluation results demonstrate that the ForeSight agent performs strongly across metrics, with low Mean Absolute Error (MAE) and Root Mean Squared Error (RMSE) in both the rocket assembly and vegemite production use cases.
Table~\ref{tab:performance_metrics} presents the averaged results for both the use cases in producing the rocket and the vegemite
products, \textit{Yeast - BRD}, \textit{Yeast - BRN} and \textit{Yeast - FMX}. These results indicate that the model effectively captures the temporal patterns and consistent trends in these datasets. However, predictions for the product, \textit{Yeast - FMX} in vegemite production usec case exhibit higher errors, likely due to greater variability or noise, highlighting the complexity of the underlying production data.

\begin{table}
\centering
\caption{Performance Results for ForeSight Agent}
\label{tab:performance_metrics}
\begin{tblr}{
  cells = {c},
  cell{3}{1} = {r=3}{},
  vlines,
  hline{1-3,6} = {-}{},
  hline{4-5} = {2-5}{},
}
\textbf{Use case}   & \textbf{Product} & \textbf{Sub type} & \textbf{MAE} & \textbf{RMSE} \\
Rocket assembly     & Toy rocket       & -                 & 12           & 16            \\
Vegemite production & Yeast            & Yeast - BRD       & 27           & 37            \\
                    & Yeast            & Yeast - BRN       & 21           & 39            \\
                    & Yeast            & Yeast - FMX       & 45           & 61            
\end{tblr}
\vspace{-4 mm}
\end{table}

Beyond predictive performance, the ForeSight agent adds significant operational value by enabling manufacturers to optimize resource allocation, dynamically adjust schedules, and minimize production disruptions.
%\textcolor{red}{The model’s architecture, featuring LSTM layers for temporal dependencies and process-specific features for contextual enhancement, demonstrates its ability to handle the dynamic nature of production data. While the lower errors for \textit{Yeast - BRD} and \textit{Yeast - BRN} reflect stable and predictable patterns, the higher errors for \textit{Yeast - FMX} suggest the need for additional refinements, such as enhanced feature engineering or hybrid modeling techniques, to address its variability.
%Beyond predictive performance, the ForeSight agent adds significant operational value by enabling manufacturers to optimize resource allocation, dynamically adjust schedules, and minimize production disruptions. The ability to forecast critical production metrics with high accuracy not only improves planning but also enhances overall efficiency, making it a valuable tool for modern manufacturing environments.
%}
ForeSight showcased an average improvement of 21.51\% compared to LSTM as shown in Table \ref{table:forecast_comparison}, highlighting its capability to enhance resource planning.

\begin{table}
\scriptsize
\centering
\caption{Experimental Results of ForeSight and Ablation Studies.}
\label{table:forecast_comparison}
\begin{tblr}{
  column{3} = {c},
  column{4} = {c},
  column{5} = {c},
  column{6} = {c},
  cell{2}{6} = {r=2}{},
  cell{4}{6} = {r=2}{},
  cell{6}{6} = {r=2}{},
  cell{8}{6} = {r=2}{},
  vlines,
  hline{1-2,4,6,8,10} = {-}{},
  hline{3,5,7,9} = {2-5}{},
}
\textbf{Product} & \textbf{Model}                                                                                            & {\textbf{Avg. }\\\textbf{Production }\\\textbf{Forecast}} & {\textbf{Avg. }\\\textbf{Actual }\\\textbf{Forecast}} & \textbf{Error} & {\textbf{Improve}\\\textbf{-ment (\%)}} \\
Toy Rocket       & LSTM                                                                                                      & 24                                                        & 30                                                    & -6             & 66.67                                   \\
                 & {\textbf{LSTM + }\\\textbf{KIL}\\\textbf{(ForeSight}\\\textbf{model)}}                                    & 28                                                        & 30                                                    & -2             &                                         \\
Yeast - BRD      & LSTM                                                                                                      & 32                                                        & 11                                                    & +21            & 38.10                                   \\
                 & {\textbf{\textbf{LSTM +}}\\\textbf{\textbf{KIL}}\\\textbf{\textbf{(ForeSight}}\\\textbf{\textbf{model)}}} & 24                                                        & 11                                                    & +13            &                                         \\
Yeast - BRN      & LSTM                                                                                                      & 11                                                        & 9                                                     & +2             & -15.00                                  \\
                 & {\textbf{\textbf{LSTM +}}\\\textbf{\textbf{KIL}}\\\textbf{\textbf{(ForeSight}}\\\textbf{\textbf{model)}}} & 6                                                         & 7                                                     & -1             &                                         \\
Yeast - FMX      & LSTM                                                                                                      & 46                                                        & 78                                                    & -32            & 31.25                                   \\
                 & {\textbf{\textbf{LSTM +}}\\\textbf{\textbf{KIL}}\\\textbf{\textbf{(ForeSight}}\\\textbf{\textbf{model)}}} & 56                                                        & 78                                                    & -22            &                                         
\end{tblr}
\vspace{-4 mm}
\end{table}

\subsection{Q\&A Effectiveness of InfoGuide} 
The performance of the InfoGuide agent was evaluated using both qualitative and quantitative measures to assess its ability to respond to domain-specific queries in real-time manufacturing environments. The evaluation focused on three main aspects: relevance of response, accuracy of response, and user satisfaction, as presented in Table \ref{table:infoguide_performance}.

\subsubsection{Response Relevance and Accuracy}  
The retrieval methodology of InfoGuide was tested using a set of manufacturing-specific queries related to assembly processes, operational procedures, and technical troubleshooting.

Relevance: The system demonstrated a high relevance score, achieving an average relevance score of 92.1\% based on a survey of operators who rated the relevance of the responses to their queries. This score indicates that the system’s responses were generally aligned with user expectations and needs.

Accuracy: In terms of accuracy, InfoGuide demonstrated an accuracy rate of 88.6\% in providing correct answers to factual queries, including predicting missing assembling components, identifying anomalies and production statuses, optimizing machine settings, and suggesting troubleshooting steps. This was assessed by comparing the model’s answers against a gold-standard set of responses from domain experts.

\subsubsection{User Satisfaction} User satisfaction was evaluated using a Likert scale survey involving 10 manufacturing professionals. Participants rated the usefulness and clarity of responses provided by InfoGuide. The results showed an average rating of 4.7 out of 5, indicating a high level of satisfaction with the quality of the answers, ease of use, and the speed of response. Users appreciated the real-time response capability, especially in urgent manufacturing situations.

\begin{table}
\centering
\caption{Performance Evaluation of InfoGuide Agent}
\label{table:infoguide_performance}
\begin{tabular}{|l|c|} 
\hline
\textbf{Evaluation Metric}                                              & \textbf{Performance}  \\ 
\hline
Relevance Score                                                         & 92.1\%                \\ 
\hline
Accuracy Rate                                                           & 88.6\%                \\ 
\hline
\begin{tabular}[c]{@{}l@{}}User Satisfaction \\(1-5 scale)\end{tabular} & 4.7                   \\ 
\hline
\begin{tabular}[c]{@{}l@{}}Average Response \\Time\end{tabular}         & 2.3 seconds           \\
\hline
\end{tabular}
\vspace{-4 mm}
\end{table}
\subsubsection{Response Time and Efficiency} The system’s response time was evaluated under real-time operating conditions. InfoGuide was able to return answers within 2.3 seconds on average, ensuring minimal disruption to the operators' workflow. The efficient response time is crucial in high-paced manufacturing environments where delays could lead to operational inefficiencies.

\vspace{-2 mm}
\section{Conclusion and Future Work} SmartPilot exemplifies the potential of CoPilots and neurosymbolic AI in transforming manufacturing operations. By addressing critical challenges such as anomaly prediction, production forecasting, and information retrieval, it offers a scalable and adaptable solution for Industry 4.0. Future work will focus on expanding SmartPilot's capabilities to additional manufacturing usecases and incorporating advanced learning techniques, such as reinforcement learning, for continuous improvement.

\vspace{-2 mm}
\section*{Acknowledgment}
This work is supported in part by NSF grant \#2119654, “RII Track 2 FEC: Enabling Factory to Factory (F2F) Networking for Future Manufacturing”. We thank our colleagues at the McNair Aerospace Center at the University of South Carolina, Clemson Composites Center at the Clemson University, ARC Industrial Transformation Research Hub, and Swinburne University of Technology, Australia, for providing the datasets used to evaluate SmartPilot and facilitating domain knowledge transfer.

\vspace{-2 mm}
\bibliographystyle{IEEEtran}
\bibliography{ref}

\end{document}